\documentclass{article}
\usepackage{spconf,amsmath,graphicx,hyperref}
\usepackage{amssymb}
\usepackage{xcolor}
\usepackage{booktabs} 
\usepackage{multirow}
\usepackage{array}
\usepackage{colortbl}

\usepackage{tikz}
\definecolor{bestcolor}{RGB}{34,139,34}
\definecolor{goodcolor}{RGB}{70,130,180}

\newcolumntype{L}[1]{>{\raggedright\arraybackslash}p{#1}}
\title{UniPACT: A Multimodal Framework for Prognostic Question Answering on Raw ECG and Structured EHR}

\name{Jialu Tang$^{1}$, Tong Xia$^{2}$, Yuan Lu$^{1}$,
Aaqib Saeed$^{1,3}$}
\address{
    $^{1}$ Eindhoven University of Technology, The Netherlands\\
    $^{2}$ Tsinghua University, China.\\
    $^{3}$ Eindhoven Artificial Intelligence Systems Institute, The Netherlands\\
    \{j.tang, y.lu, a.saeed\}@tue.nl, tongxia@mail.tsinghua.edu.cn
}
\begin{document}
\maketitle

\begin{abstract}
Accurate clinical prognosis requires synthesizing structured Electronic Health Records (EHRs) with real-time physiological signals like the Electrocardiogram (ECG). Large Language Models (LLMs) offer a powerful reasoning engine for this task but struggle to natively process these heterogeneous, non-textual data types. To address this, we propose UniPACT (Unified Prognostic Question Answering for Clinical Time-series), a unified framework for prognostic question answering that bridges this modality gap. UniPACT's core contribution is a structured prompting mechanism that converts numerical EHR data into semantically rich text. This textualized patient context is then fused with representations learned directly from raw ECG waveforms, enabling an LLM to reason over both modalities holistically. We evaluate UniPACT on the comprehensive MDS-ED benchmark, it achieves a state-of-the-art mean AUROC of 89.37\% across a diverse set of prognostic tasks including diagnosis, deterioration, ICU admission, and mortality, outperforming specialized baselines. Further analysis demonstrates that our multimodal, multi-task approach is critical for performance and provides robustness in missing data scenarios.
\end{abstract}

\section{Introduction}
\begin{figure*}[t]
\centering
\includegraphics[width=0.95\textwidth]{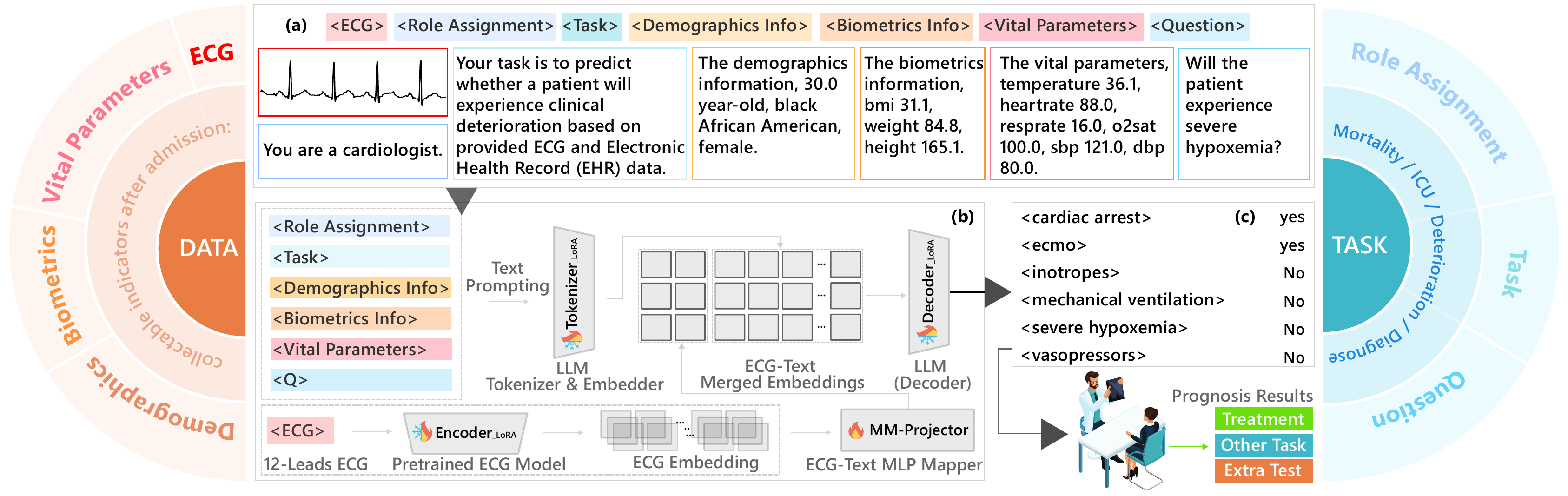}
\caption{\small{The UniPACT framework for multimodal prognostic question answering. 
\textbf{(a) Structured Prompt Formulation:} Heterogeneous patient data, including structured EHR (demographics, biometrics, vitals) and a reference to the ECG waveform, are converted into a unified natural language prompt. This process transforms numerical values into a format that is natively understandable by the LLM.
\textbf{(b) Multimodal Fusion Architecture:} A pretrained encoder processes the raw 12-lead ECG waveform to produce a feature embedding. A multimodal projector (MM-Projector) then aligns this ECG embedding with the LLM's text embedding space. These aligned ECG features are seamlessly integrated with the tokenized text prompt and processed by the LLM decoder for unified reasoning.
\textbf{(c) Prognostic Output Generation:} The model generates a direct answer to the prognostic question (e.g., `Yes'/`No' to a query about clinical deterioration).
}}
\label{fig:main}
\end{figure*}

Accurate patient prognosis in acute care is a cornerstone of modern medicine, directly guiding critical decisions such as ICU admission, treatment selection, and risk intervention  \cite{smit2023causal}. Clinicians formulate a prognosis not from a single data point, but by synthesizing a holistic view of the patient: their static baseline (demographics, comorbidities from EHR), dynamic state (vital signs), and acute physiological signals (like the 12-lead Electrocardiogram) \cite{raghunath2020prediction}. A subtle T-wave inversion (from the ECG), for instance, may signify a critical mortality risk, but only when contextualized by abnormal potassium levels and a history of hypertension (from the EHR). Capturing this complex, cross-modal reasoning is the central challenge in computational prognosis.

For decades, this challenge was met with simplified clinical risk scores, which are limited by manual feature selection and linear assumptions. While standard machine learning models offered improved accuracy, they are predominantly designed as rigid, single-task predictors~\cite{alcaraz2407mds}. These models lack the flexibility to adapt to the diverse, dynamic questions clinicians face. They cannot be naturally queried about different outcomes (e.g., diagnosis, deterioration, mortality) and fundamentally struggle to fuse the dense, continuous language of ECG signals with the discrete, numerical language of structured EHR tables.

The advent of Large Language Models (LLMs) introduces a new paradigm \cite{li2023llava,singhal2023large, li2024mediq}, offering a path from rigid prediction to flexible, prompt-based prognostic \textit{question answering}. The vision is a single, unified model that a clinician can query in natural language: ``Given this patient's history and their current ECG, what is their risk of severe hypoxemia?'' However, this vision is blocked by a fundamental barrier: LLMs are text-native reasoning engines. They cannot natively interpret the most critical prognostic data. Raw ECG waveforms are dense signals where subtle morphology holds diagnostic meaning, while structured EHR data consists of context-poor numerical values. Current workarounds, such as summarizing ECGs into text reports \cite{yu2024ecg, xu2024penetrative}, result in critical information loss, discarding the very waveform patterns essential for expert-level prognosis.

To bridge this gap and enable better prognostic reasoning, we propose UniPACT (Unified Prognostic Question Answering for Clinical Time-series). UniPACT is a unified framework designed for multimodal prognostic question answering and makes raw physiological signals and structured data ``legible'' to an LLM. For the ECG, it employs a dedicated waveform encoder to learn deep representations directly from the 12-lead ECG signal. For the EHR, it uses a novel structured prompting mechanism to convert numerical data into semantically rich sentences (e.g., ``The patient's heart rate is 88 beats per minute''), embedding them with vital clinical context. This strategy enables the LLM to simultaneously process high-fidelity ECG signals and structured EHR data. By framing diverse prognostic queries within a single question-answering framework, UniPACT can seamlessly switch between predicting different outcomes from long-term mortality to immediate clinical deterioration without requiring separate models. Our work makes following contributions:
\begin{itemize}
    \item We introduce UniPact, the first framework to unify raw ECG waveforms with structured EHR data, overcoming the modality bottleneck in LLM-based prognostic reasoning.
    \item We propose Structured EHR Prompting as a key design choice in multimodal prognosis. This mechanism provides an effective and flexible way that jointly unifies raw ECG signals and heterogeneous numerical clinical data while preserving both numerical precision and semantics. This design underpins the medical relevance of our approach by enabling clinically grounded question answering in a wide spectrum of queries.
    \item Through comprehensive evaluation on the MDS-ED~\cite{alcaraz2407mds}
    (MIMIC-IV-ECG \& MIMIC-IV derived) benchmark, we demonstrate that UniPact significantly outperforms established baselines and maintains robustness in both multi-task and missing-modality scenarios.
\end{itemize}

\section{Method}
\label{sec:method}
The UniPACT framework is designed to perform prognostic question answering by unifying raw ECG waveforms and structured EHR data within an end-to-end generative model. As illustrated in Figure~\ref{fig:main}, our architecture consists of three core components: (1) a dedicated encoder for raw ECG signals, (2) a structured prompting mechanism to textualize EHR data, and (3) a large language model (LLM) that fuses these multimodal representations to generate a final prediction.

\subsection{Modality-Specific Representation Learning}

\textbf{ECG Waveform Encoder.} To capture the rich diagnostic information in physiological signals, we process raw 12-lead ECG waveforms directly, avoiding lossy conversion to text reports. We employ the pre-trained Transformer-based ECG encoder from D-BETA~\cite{pham2024c}, which has been shown to be effective at learning discriminative representations from ECG signals. Given a raw ECG signal $E \in \mathbb{R}^{L \times C}$ (where $L$ is the number of time steps, typically 5000 for a 10-second recording at 500 Hz, and $C=12$ leads), the encoder outputs a sequence of feature embeddings:
\begin{equation}
    H_{ecg} = \text{ECG-Encoder}(E),
\end{equation}
where $H_{ecg} \in \mathbb{R}^{N \times d_{ecg}}$ is a sequence of $N$ embedding vectors, each of dimension $d_{ecg}$. This approach preserves the fine-grained temporal and morphological patterns of the waveform.

\noindent \textbf{Structured EHR Prompting.} 
A key novelty of UniPACT is its method for making structured EHR data comprehensible to an LLM. Rather than using raw values, we convert $<Demographics>$ (3 parameters), $<Biometrics>$ (3 parameters), and $<Vital\ Parameters>$ (7 parameters) into natural language sentences via predefined templates. This representation, $T_{\text{EHR}}$, maintains numerical precision while providing the contextual cues LLMs require. An example prompt follows:
\begin{flushleft}
\small
\small{\texttt{The demographics information: 30.0 year-old, black African American, female. The vital parameters: temperature 36.1, heartrate 88.0, resprate 16.0. The biometrics information: bmi 31.1, weight 84.8, height 165.1.}}
\end{flushleft}

\subsection{Multimodal Fusion and Generation}
\textbf{Embedding Space Alignment.} To fuse these diverse modalities, we first align their representations within the LLM's embedding space. The ECG feature embeddings $H_{ecg}$ are mapped into the LLM's word embedding dimension $d_{llm}$ using a small, trainable projection network, which we implement as a two-layer MLP (the MM-Projector):
\begin{equation}
    H'_{ecg} = \text{MLP}(H_{ecg}),
\end{equation}
where $H'_{ecg} \in \mathbb{R}^{N \times d_{llm}}$. The textualized EHR prompt, $T_{\text{EHR}}$, is tokenized and embedded using the LLM's native tokenizer and embedding layer, resulting in embeddings $H_{\text{EHR}} \in \mathbb{R}^{M \times d_{llm}}$.

\noindent \textbf{Unified Input and Autoregressive Prediction.} The final input to the LLM is a single, unified sequence constructed by concatenating the processed multimodal embeddings. We base our model on the LLaVA framework~\cite{liu2023visual} and use MedGemma-4B~\cite{sellergren2025medgemma} as the backbone LLM as it is pretrained on medical data. The complete input sequence is formatted as:
\begin{equation}
    H_{\text{input}} = [ 
    H'_{ecg},  H_{\text{prompt}},  H_{\text{question}} ],
\end{equation}
where $H_{\text{prompt}}$ contains the embedded EHR data and task instructions, and $H_{\text{question}}$ is the embedded prognostic query (e.g., ``Will the patient experience severe hypoxemia?''). The LLM then autoregressively predicts the answer $Y$ (e.g., ``Yes'' or ``No'') based on this fused multimodal context.

\subsection{Multi-Task Learning via Unified Prompting}
UniPACT is trained as a single model on a diverse set of prognostic tasks (diagnosis, deterioration, ICU admission, mortality). We unify all tasks using a consistent prompt structure, which instructs the model on its role, the specific task, and the question to answer. A generalized template is as follows:
\begin{quote}
\small
\texttt{<ECG EMBEDDINGS>} \\
\texttt{<Role Assignment> <Task Description>} \\
\texttt{<EHR Information as Text>} \\
\texttt{<Specific Prognostic Question>?} \\
\texttt{Answer strictly with Yes or No.}
\end{quote}
The entire model is trained end-to-end using a standard language modeling objective, which maximizes the likelihood of the ground-truth answer tokens. Specifically, we minimize the cross-entropy loss only on the answer portion of the sequence:
\begin{equation}
    \mathcal{L} = - \sum_{i=1}^{k} \log P(y_i | H_{\text{input}}, y_{<i}; \theta),
\end{equation}
where $Y = (y_1, ..., y_k)$ are the tokens of the target answer (e.g., ``Yes''), and $\theta$ represents the model parameters. 
To fully leverage information from ECG and textual modalities and ensure training efficiency, we trained in two stages: in the first stage, we keep the ECG encoder and LLM weights frozen, updating only the MM-Projector. We employed LoRA~\cite{hu2022lora} fine-tuning for the ECG encoder to reduce the number of trainable parameters. In the second stage, we inserted LoRA adapters (rank $r$=128, scaling $\alpha$=256, dropout=0.05) to the linear layers in the ECG encoder and LLM.

\section{Results}
\label{sec:results}
We evaluate UniPACT's performance on a comprehensive suite of prognostic tasks from the MDS-ED benchmark~\cite{alcaraz2407mds} derived from MIMIC-IV-ECG~\cite{gow2023mimic} \& MIMIC-IV~\cite{johnson2023mimic} dataset. Our analysis focuses on four key aspects: (1) comparison against state-of-the-art baselines; (2) the contribution of individual modalities; (3) the value of our unified multi-task learning approach; and (4) an exploratory comparison against general-purpose LLM APIs. We use the Area Under the Receiver Operating Characteristic Curve (AUROC) as the primary evaluation metric across all prognostic tasks.\\

\noindent \textbf{Comparison with Baseline Methods.} We begin with comparing UniPACT's performance against established models on the considered benchmark in Table~\ref{tab:main_results}. 
The baselines include two recent ECG–language models, ECG-Chat~\cite{zhao2024ecg} and Q-HEART~\cite{pham2025qheart}, as well as a specialized multimodal classification model, MDS-ED~\cite{alcaraz2407mds}. The latter comprises two independently trained models: one for Deterioration, and another for Deterioration, ICU, and Mortality. It represents the prior state-of-the-art on this dataset. UniPACT achieves an overall AUROC of 89.37\%, outperforming all baselines. Notably, it surpasses the highly specialized MDS-ED model, highlighting the strength of our framework. For fair comparison, we use same metrics as MDS-ED. In Table~\ref{tab:main_results} the parenthetical numbers shows the count of individual sub-tasks (out of 1443) where a model achieves robust performance (AUROC 95\% CI lower bound $>$ 0.8). UniPACT demonstrates strong performance on 883 sub-tasks, a significant increase from the 623 achieved by the MDS-ED, indicating greater reliability across a wider range of clinical scenarios. The general-purpose models underperformed, underscoring the importance of fine-tuning for these specific prognostic tasks.

\begin{table}[t]
\centering
\caption{\small{Comparative Performance Analysis on Cardiovascular Prognostic Tasks.}}
\label{tab:main_results}
\resizebox{\columnwidth}{!}{%
\begin{tabular}{@{}l*{4}{c}c@{}}
\toprule
& \multicolumn{4}{c}{\textbf{Task-Specific Performance (AUROC \%)}} & \\
\cmidrule(lr){2-5}
\textbf{Method} & \textbf{Diagnosis} & \textbf{Deterioration} & \textbf{ICU} & \textbf{Mortality} & \textbf{Overall} \\
& \footnotesize{(1428 classes)} & \footnotesize{(6 classes)} & \footnotesize{(2 classes)} & \footnotesize{(7 classes)} & \footnotesize{(1443 total)} \\
\midrule
\multicolumn{6}{@{}l}{\textit{ECG-based Foundation Models}} \\
\midrule
ECG-Chat \cite{zhao2024ecg} & 49.70 & 57.54 & 56.40 & 55.78 & 54.86 \\
& \footnotesize{53/1428} & \footnotesize{0/6} & \footnotesize{0/2} & \footnotesize{0/7} & \footnotesize{53/1443} \\
Q-HEART \cite{pham2025qheart} & 50.15 & 55.42 & 54.89 & 56.32 & 54.20 \\
& \footnotesize{139/1428} & \footnotesize{0/6} & \footnotesize{0/2} & \footnotesize{0/7} & \footnotesize{139/1443} \\
\midrule
\multicolumn{6}{@{}l}{\textit{Multimodal Approaches}} \\
\midrule
MDS-ED~\cite{alcaraz2407mds} & 82.56 & 90.70 & 90.63 & 91.68 & 88.90 \\
& \footnotesize{609/1428} & \footnotesize{5/6} & \footnotesize{2/2} & \footnotesize{7/7} & \footnotesize{623/1443} \\
\rowcolor{gray!10}
\textbf{UniPACT (ours)} & \textcolor{bestcolor}{\textbf{83.98}} & \textcolor{bestcolor}{\textbf{91.17}} & \textbf{90.50} & \textcolor{bestcolor}{\textbf{91.82}} & \textcolor{bestcolor}{\textbf{89.37}} \\
\rowcolor{gray!10}
& \footnotesize{\textbf{868/1428}} & \footnotesize{\textbf{6/6}} & \footnotesize{\textbf{2/2}} & \footnotesize{\textbf{7/7}} & \footnotesize{\textbf{883/1443}} \\
\midrule
\multicolumn{6}{@{}l}{\textit{Relative Improvement (\%)}} \\
\midrule
$\Delta$ vs. ECG-Chat & +69.0\% & +58.4\% & +60.5\% & +64.6\% & +62.9\% \\
$\Delta$ vs. MDS-ED & +1.7\% & +0.5\% & -0.1\% & +0.2\% & +0.5\% \\
\bottomrule
\end{tabular}
}
\end{table}

\begin{table}[t]
\centering
\caption{\small{Comparison with Large Language Model APIs.} \small{$^\dagger$Using zero-shot prompting with ECG reports and diagnosis All values in AUROC (\%).}}
\label{tab:llm_comparison}
\resizebox{\columnwidth}{!}{%
\begin{tabular}{@{}l*{5}{c}@{}}
\toprule
\textbf{Method} & \textbf{Diagnosis} & \textbf{Deterioration} & \textbf{ICU} & \textbf{Mortality} & \textbf{Overall} \\
\midrule
GPT-5-chat$^\dagger$ & 55.68 & 78.12 & 70.04 & 62.27 & 66.53 \\
Gemini-2.5 Pro$^\dagger$ & 57.82 & 78.72 & 70.70 & 72.38 & 69.41 \\
\midrule
\textbf{UniPACT (ours)} & \textbf{83.98} & \textbf{91.17} & \textbf{90.50} & \textbf{91.82} & \textbf{89.37} \\
\midrule
\multicolumn{6}{@{}l}{\textit{Performance Gap}} \\
vs. GPT-5-chat & +28.30 & +13.05 & +20.46 & +29.55 & +22.84 \\
vs. Gemini & +26.16 & +12.45 & +19.80 & +19.44 & +19.96 \\
\bottomrule
\end{tabular}
}
\end{table}

\noindent \textbf{Exploratory Comparison with LLM APIs.} We conducted an exploratory study to benchmark UniPACT against powerful, proprietary LLMs like GPT-5-Chat and Gemini-2.5 Pro, as shown in Table~\ref{tab:llm_comparison}. It is crucial to note that this is not a direct, apples-to-apples comparison. While UniPACT is fine-tuned on the task-specific data and processes raw ECG waveforms, the LLM APIs were prompted in a zero-shot manner and were given a \textit{textual description and diagnoses} of the ECG findings instead of the raw signal itself. For evaluation, we sampled 20,000 instances from ~400 tasks, maintaining a balanced positive and negative cases to cover diverse clinical conditions while keeping API cost into account. Under this setup, UniPACT significantly outperforms the general-purpose APIs. This result is not intended to be a critique of these powerful models, but rather to highlight a key finding: for complex, domain-specific tasks like clinical prognosis, the ability to process raw modal data (like ECG signals) and fine-tune on relevant data is critical for achieving better performance. 


\begin{table}[t]
\centering
\scriptsize
\caption{\small{Comprehensive Ablation Analysis of UniPACT.}}
\label{tab:ablation_comprehensive}
\resizebox{\columnwidth}{!}{%
\begin{tabular}{@{}L{3.5cm}*{5}{r}@{}}
\toprule
\multirow{2}{*}{\textbf{Model Configuration}} & \multicolumn{5}{c}{\textbf{Performance Metrics (AUROC \%)}} \\
\cmidrule(l){2-6}
& Diagnose & Deterioration & ICU & Mortality & Mean$\pm$SD \\
\midrule
\multicolumn{6}{@{}l}{\textbf{A. Modality Analysis}} \\
\midrule
ECG-based & 66.05 & 74.80 & 71.89 & 79.79 & 73.13$\pm$5.74 \\
EHR-based & 69.97 & 87.36 & 82.62 & 83.37 & 80.83$\pm$7.53 \\
\textbf{UniPACT (Multimodal)} & \textbf{83.98} & \textbf{91.17} & \textbf{90.50} & \textbf{91.82} & \textbf{89.37$\pm$3.63} \\
\quad $\Delta$ vs ECG & +17.93 & +16.37 & +18.61 & +12.03 & +16.24 \\
\quad $\Delta$ vs EHR & +14.01 & +3.81 & +7.88 & +8.45 & +8.54 \\
\midrule
\multicolumn{6}{@{}l}{\textbf{B. Learning Paradigm}} \\
\midrule
Single-Task Learning & \textbf{84.16} & 90.15 & 81.94 & 90.26 & 86.63$\pm$4.23 \\
\textbf{Multi-Task Learning} & 83.98 & \textbf{91.17} & \textbf{90.50} & \textbf{91.82} & \textbf{89.37$\pm$3.63} \\
\quad $\Delta$ (MTL gain) & $-$0.18 & +1.02 & +8.56 & +1.56 & +2.74 \\
\midrule
\multicolumn{6}{@{}l}{\textbf{C. Feature Ablation} (Performance when component is removed)} \\
\midrule
w/o Demographics & 74.82 & 80.75 & 80.91 & 81.29 & 79.44$\pm$3.09 \\
w/o Biometrics & 77.65 & 85.12 & 84.01 & 85.62 & 83.10$\pm$3.70 \\
w/o Vitals & 72.12 & 78.89 & 77.95 & 79.32 & 77.07$\pm$3.35 \\
\textbf{w/o ECG} & \textbf{50.38} & \textbf{54.27} & \textbf{54.43} & \textbf{54.60} & \textbf{53.42$\pm$2.03} \\
w/o EHR & 73.01 & 79.89 & 78.95 & 80.31 & 78.04$\pm$3.40 \\
\textit{Full Model} & \textit{83.98} & \textit{91.17} & \textit{90.50} & \textit{91.82} & \textit{89.37$\pm$3.63} \\
\bottomrule
\end{tabular}
}
\end{table}

\noindent \textbf{Role of Multimodality.} To quantify the benefit of integrating ECG and EHR data, we evaluated uni-modal versions of UniPACT. Table~\ref{tab:ablation_comprehensive} (A) presents the performance of models trained with only ECG or only EHR data compared to the full multimodal UniPACT. Our results clearly demonstrate strong synergistic effects. The EHR-only model achieves a 80.83\% AUROC, confirming that structured clinical data is highly predictive. However, the full UniPACT model, which integrates raw ECG waveforms, improves performance by a substantial margin of +8.54\% AUROC. This gain confirms our central hypothesis: UniPACT effectively leverages the unique, complementary information present in both the patient's clinical history (EHR) and their real-time physiological state (ECG) to form a more complete and accurate prognostic assessment.

\noindent \textbf{Benefit of Multi-Task Learning.} Our framework trains a single, unified model for all prognostic tasks. To validate this approach, we compared our multi-task UniPACT model against single-task counterparts, where a separate model was trained for each of the four main task categories. As shown in Table~\ref{tab:ablation_comprehensive} (B), the unified multi-task model achieves a higher overall AUROC (89.37\% vs. 86.63\%). The performance lift, particularly in the ICU and Mortality tasks, suggests that the model learns shared, generalizable representations of patient state that are beneficial across different prognostic horizons.

\noindent \textbf{Robustness to Missing Data.} In clinical practice, patient data is often incomplete. We assessed UniPACT's robustness by evaluating its performance when specific components of the EHR are absent. Table~\ref{tab:ablation_comprehensive} (C) presents the impact of removing Demographics, Biometrics, or Vitals. Our results show that while every data component contributes to the final prediction, the model exhibits graceful degradation rather than catastrophic failure. For example, removing patient vitals—a highly informative feature set—reduces the overall AUROC from 89.37\% to 77.07\%. While this is a significant drop, the resulting performance is still substantially better than random chance and superior to the uni-modal ECG model, indicating that UniPACT effectively re-weights the available information (in this case, ECG, demographics, and biometrics) to compensate for the missing data. 

\section{Conclusion}
We presented UniPACT, a unified framework that effectively integrates raw ECG waveforms and structured EHR data for LLM-based prognosis. By translating numerical EHR into semantic prompts and fusing them with deep ECG features, our model achieves superior performance on the MDS-ED benchmark, outperforming established baselines. The results demonstrate that a single, multi-task generative model can surpass specialized systems in both accuracy and robustness. Our approach enhances the ability of LLMs to reason over complex clinical data by providing high-fidelity, multimodal inputs.

\vfill\pagebreak

\section{Acknowledgments and Ethical Compliance}
No funding was received for conducting this study. The authors declare that they have no relevant financial or nonfinancial interests to disclose. This study made use of publicly available and fully anonymized human data sets. In accordance with the policies of the data providers, no additional institutional review board approval was required.

\bibliographystyle{IEEEbib}
\bibliography{main_ref}

\end{document}